%% file: main.tex
\documentclass{article}
\usepackage{titling}
\usepackage{url}
\usepackage{amsmath,amsthm,verbatim,amssymb,amsfonts,amscd, graphicx, enumitem}
\usepackage{graphics}
\usepackage{centernot}
\usepackage{authblk}
\usepackage{bm}
\usepackage[ruled]{algorithm2e}
\usepackage{booktabs}
\usepackage{xcolor}

\usepackage{algpseudocode}

\usepackage[colorlinks,linkcolor=blue,citecolor=blue]{hyperref}

\usepackage[bottom]{footmisc}
\usepackage{caption}
\usepackage{subcaption}
\usepackage{helvet}  
\usepackage{courier}  
\usepackage{url}  
\usepackage{graphicx}  
\usepackage{multirow}
\usepackage{color}
\usepackage{MnSymbol}
\usepackage{makecell}
\usepackage[dvipsnames]{xcolor}
\usepackage{caption} 
\usepackage{natbib}
\usepackage{bbm}

\usepackage{textcomp}
\usepackage{wrapfig}

\usepackage{csquotes}
\usepackage{cleveref}

\newtheorem{theorem}{Theorem}
\newtheorem{lemma}{Lemma}
\newtheorem{corollary}{Corollary}

\newtheorem{definition}{Definition}

\crefformat{manualeq}{eq.\,(#2#1#3)}
\crefmultiformat{manualeq}
  {eqs.\,(#2#1#3)} 
  { and (#2#1#3)}
  { and (#2#1#3)}
  { and (#2#1#3)}
\crefrangeformat{manualeq}{eqs.\,(#3#1#4)--(#5#2#6)}

\newcounter{manualeq}
\newenvironment{myeqwrapper}[1]{%
  \renewcommand{\themanualeq}{#1}
  \refstepcounter{manualeq}
}{}
\crefname{manualeq}{eq.}{eqs.}
\Crefname{manualeq}{Eq.}{Eqs.}

\crefname{equation}{eq.}{eqs.}
\Crefname{equation}{Eq.}{Eqs.}

\definecolor{color1}{HTML}{4F46E5}
\definecolor{color2}{HTML}{0F763E}

\title{{\fontsize{20pt}{24pt}\selectfont An Equivariance Toolbox for Learning Dynamics}}

\author{
\textbf{Yongyi Yang}$^{1,3}$, \textbf{Liu Ziyin}$^{2,3}$ \\
\begin{flushleft}
\hspace{5.5em}$^1$ University of Michigan \\
\hspace{5.5em}$^2$ Massachusetts Institute of Technology \\
\hspace{5.5em}$^3$ NTT Research
\end{flushleft}}

\date{}
\predate{} 
\postdate{} 

\begin{document}

\maketitle

\begin{abstract}
\noindent
Many theoretical results in deep learning can be traced to symmetry or equivariance of neural networks under parameter transformations. However, existing analyses are typically problem-specific and focus on first-order consequences such as conservation laws, while the implications for second-order structure remain less understood. We develop a general equivariance toolbox that yields coupled first- and second-order constraints on learning dynamics. The framework extends classical Noether-type analyses in three directions: from gradient constraints to Hessian constraints, from symmetry to general equivariance, and from continuous to discrete transformations. At the first order, our framework unifies conservation laws and implicit-bias relations as special cases of a single identity. At the second order, it provides structural predictions about curvature: which directions are flat or sharp, how the gradient aligns with Hessian eigenspaces, and how the loss landscape geometry reflects the underlying transformation structure. We illustrate the framework through several applications, recovering known results while also deriving new characterizations that connect transformation structure to modern empirical observations about optimization geometry.

\end{abstract}

\section{Introduction}

A large body of recent theory in deep learning exploits, either explicitly or implicitly, \emph{symmetry} or \emph{equivariance} of neural networks with respect to parameter transformations \citep{hidenori_symmetry,noether,kaifeng_homo,relu_balance,vidal_deep_linear}. 
Such structures arise naturally in many settings: layer-wise rescaling in ReLU networks, reparameterizations in deep linear models, and various invariances induced by over-parameterization. 
While powerful, existing analyses are typically developed in a problem-specific manner: one identifies a particular transformation, derives a tailored identity for a specific architecture or loss, and builds a dedicated argument around that identity. 
As a result, the \emph{common mechanism} underlying these diverse results remains obscured, and extensions beyond specific first-order identities are often unclear.

A related line of work applies Noether's theorem to systematize symmetry-based arguments \citep{noether,bozhao_conserved,hidenori_symmetry}. 
This perspective elegantly derives conservation laws from continuous symmetries, but it is intrinsically a \emph{first-order} tool: it constrains the gradient but says little about curvature. 
Yet many actively studied phenomena concern \emph{second-order} structure such as the Hessian spectrum, the geometry of sharp and flat directions, and the low-dimensional subspaces that dominate optimization \citep{eigval_of_hessian,eos,gd_tiny_space,sam}. 
Although \citet{hidenori_symmetry} derived some second-order identities for certain symmetries, these were not developed into a general characterization, and the scope was limited to continuous symmetry rather than general equivariance or discrete symmetry.

In this work, we develop a unified \emph{equivariance toolbox} for learning dynamics that yields coupled first- and second-order constraints. 
Our framework applies whenever a model $f$ satisfies an equivariance relation: transforming parameters by an operation $H$ is equivalent to transforming outputs by an operation $G$. 
From this relation, we derive universal identities that constrain both the gradient and Hessian of the loss along transformation-induced directions. Importantly, this framework can handle both continuous and discrete equivariances (symmetries). 
When the equivariance is continuous (i.e., continuously parameterized), these identities hold throughout parameter space; while when it is discrete (e.g., sign flips or permutations), weaker but still informative constraints hold on the fixed-point sets of the transformation.

At the first order, our framework recovers a broad family of known results as special cases. 
For symmetries (where $G$ is the identity), it reduces to orthogonality between the gradient and symmetry-generated directions, recovering Noether-type conservation laws. 
For non-trivial equivariances such as homogeneity, it reproduces key scalar relations that underpin classical implicit-bias arguments \citep{kaifeng_homo}. Both cases are handled by the same theoretical framework.

At the second order, our framework provides characterizations that are largely missing in existing work. 
The derived identities constrain how the Hessian acts along transformation-induced directions, yielding concrete structural predictions: relationships between the gradient and the Hessian's column space, sharpness estimates along canonical directions, and constraints on eigenspace structure. 
These predictions connect directly to empirical observations about optimization geometry and late-stage training behaviors \citep{gd_tiny_space,eos,nc,nc_review}.

We illustrate the framework through several applications. 
For homogeneous networks, the second-order identities reveal why gradient descent may concentrate in low-dimensional subspaces and how sharpness evolves during training. For network with symmetries, they provide new characterization of the Noether flow when the network is trained by stochastic gradient flow. 
For last-layer transformations, they connect parameter-space curvature to output-space geometry, shedding light on phenomena related to neural collapse. 
For discrete symmetries such as mirror symmetry, the fixed-point identities recover and generalize recent results on gradient and Hessian structure \citep{ziyin_mirror_symmetry}.

\paragraph{Organization.}
\Cref{sec:prelim} introduces tensor notations used throughout. 
\Cref{sec:main} presents the general framework and states the main identities for both continuous equivariance and discrete symmetry. 
\Cref{sec:firstorder} instantiates the first-order identity, recovering conservation laws and homogeneity-based relations. 
\Cref{sec:secondorder} develops the second-order consequences, illustrating them through homogeneity, last-layer transformations, and continuous symmetry. 
\Cref{sec:discrete} treats discrete symmetries and recovers mirror-symmetry results as a special case. 
All proofs are deferred to the appendix.

\section{Preliminaries: Tensor Operations}\label{sec:prelim}

Throughout this paper we will frequently manipulate higher-order derivatives and compositions of tensor-valued maps. To avoid ambiguity, we fix a notation system for tensors and tensor functions. Throughout, we view a tensor as a (curried) linear mapping: it takes one vector as input and outputs another tensor. This convention is consistent with the treatment in modern formal language systems such as Lean 4 \citep{lean4}.

Formally, a \textbf{tensor shape} is a finite sequence of positive integers. For shapes $s$ and $t$, we write $(s,t)$ for their concatenation. For an integer $n$, we write $(n)$ for the shape consisting of the single element $n$ (and we may omit parentheses when there is no ambiguity). For a shape $s$, we denote by $\mathbb T(s)$ the set of all tensors of shape $s$, defined recursively:
\begin{enumerate}[leftmargin=2.0em, itemsep=0.1em]
    \item A linear mapping $\mathbb R^n \to \mathbb R$ is a \textbf{tensor} of shape $(n)$.
    \item Given a shape $s$, a linear mapping $\mathbb R^n \to \mathbb T(s)$ is a \textbf{tensor} of shape $(n,s)$.
\end{enumerate}
For a shape $s$, a positive integer $n$, and a (sufficiently smooth) function $f:\mathbb R^n \to \mathbb T(s)$, then the Fr\'echet derivative (gradient) of $f$ is denoted as $\nabla f: \mathbb R^n \to \mathbb T(n ,s)$.

For functions $f$ and $g$, we use $g\circ f$ to denote their (usual) composition whenever it is well-defined. We also define a more general notion of composition, the \textbf{$k$-th component composition} $\circ_k$, which is convenient under the curried representation of tensors. We define it recursively as follows:
\begin{enumerate}[leftmargin=2.0em, itemsep=0.1em]
    \item The first component composition coincides with standard composition: $\circ_1 := \circ$.
    \item For $k\ge 2$, define $(g\circ_k f)(x) := g(x)\circ_{k-1} f$ whenever the right-hand side is well-defined.
\end{enumerate}
Intuitively, $\circ_k$ composes $f$ into the $k$-th input component of $g$ when one interprets curried tensors as functions of multiple arguments. For example, if $g \in \mathbb T(a,b,c)$ and $f \in \mathbb T(d,b)$, then $g \circ_2 f \in \mathbb T(a,d,c)$ and $(g\circ_2 f)(x,y,z) = g(x, f(y), z)$.

\paragraph{Example.} Readers familiar with matrix operations may find the following correspondence helpful. A tensor of shape $(n,m)$ can be viewed with an $n \times m$ matrix; a tensor of shape $(n)$ can be viewed as an $n$-dimensional vector. For example, if $A \in \mathbb T(n,m)$ and $B$ is another tensor (can be a matrix or a vector, as long as the dimension matches), the composition $A \circ B$ corresponds to matrix multiplication $AB$, and the second-component composition $A \circ_2 B$ corresponds to $B^\top A$. Specifically, if $x \in \mathbb T(n)$ and $y \in \mathbb T(m)$ are vectors, we have $A \circ x \circ_2 y = y^\top A x = \left<y, Ax\right>$.  

\section{Main Results: A General Equivariance Framework}\label{sec:main}

In this section we present the main theoretical results: universal first- and second-order identities induced by equivariance. We consider two cases separately. When the equivariance is continuous, the identities hold at every point in parameter space. When the transformation is not smoothly parameterized but admits fixed points, weaker constraints still hold on the fixed-point set. All statements use the tensor notations from \Cref{sec:prelim}; concrete instantiations appear in \Cref{sec:firstorder,sec:secondorder,sec:discrete}.

Throughout, we consider a model $f:\mathbb R^d \to \mathbb R^c$, representing a neural network as a function of its $d$ trainable parameters with $c$-dimensional output (input data can be absorbed into the definition of $f$). We also consider a smooth loss function $\ell:\mathbb R^c \to \mathbb R$, so that the overall objective is $L := \ell \circ f:\mathbb R^d \to \mathbb R$.

We introduce two families of transformations. The first, $H:\mathbb R^d \times \mathbb R^p \to \mathbb R^d$, acts on parameters and is parameterized by a $p$-dimensional vector ${\boldsymbol{\lambda}}\in \mathbb R^p$. We write $\nabla_\theta H$ and $\nabla_\lambda H$ for the Fr\'echet derivatives (gradients) of $H$ with respect to its first and second arguments, respectively. The second, $G:\mathbb R^c \times \mathbb R^p \to \mathbb R^c$, acts on outputs, with gradients $\nabla_y G$ and $\nabla_\lambda G$.

\begin{definition}[Equivariance and good position]\label{def:equiv-good}
Assume $f$, $H(\cdot, {\boldsymbol{\lambda}})$, and $G(\cdot, {\boldsymbol{\lambda}})$ are smooth for every ${\boldsymbol{\lambda}} \in \mathbb R^p$. We say that $f$ admits an \textbf{$(H,G)$-equivariance} if
\begin{align}
f\big(H({\boldsymbol{\theta}},{\boldsymbol{\lambda}})\big)=G\big(f({\boldsymbol{\theta}}),{\boldsymbol{\lambda}}\big)
\quad \text{for all } {\boldsymbol{\theta}}\in \mathbb R^d \text{ and } {\boldsymbol{\lambda}}\in \mathbb R^p.
\label{eq:equivariance}
\end{align}
If additionally $H$ and $G$ are both smooth in ${\boldsymbol{\lambda}}$, we say this equivariance is \textbf{continuous}. A pair $({\boldsymbol{\theta}},{\boldsymbol{\lambda}})$ is called a \textbf{good position} if both $\nabla_\theta H({\boldsymbol{\theta}},{\boldsymbol{\lambda}})$ and $\nabla_y G(f({\boldsymbol{\theta}}),{\boldsymbol{\lambda}})$ are invertible.
\end{definition}

An important special case of equivariance is \emph{symmetry} (or \emph{invariance}), where $G(\cdot, {\boldsymbol{\lambda}}) = \mathrm{Id}$ is the identity mapping. In some physics-related contexts, the set of all symmetries is required to form a group \citep{bozhao_conserved,bo_zhao_symmetry,hidenori_symmetry}. While this is often the case in practice, our framework does not require any algebraic structure on the symmetric or equivariant transformations. We only require smoothness near the good position, which is a purely analytic condition.

Below, we present two results addressing continuous and discrete equivariances separately. We emphasize that both theorems arise from the same equivariance condition and share a common derivation strategy (see \Cref{sec:proof-of-main} for details).

\subsection{Continuous Equivariance}

We first present results for continuous equivariances, where the transformation parameter ${\boldsymbol{\lambda}}$ varies continuously. 

\begin{definition}[Characteristic direction and characteristic output]\label{def:charXY}
Suppose $f$ admits a continuous $(H,G)$-equivariance and $({\boldsymbol{\theta}},{\boldsymbol{\lambda}})$ is a good position. Define
\begin{align*}
\begin{cases}
X({\boldsymbol{\theta}},{\boldsymbol{\lambda}})
= \nabla_\theta H({\boldsymbol{\theta}},{\boldsymbol{\lambda}})^{-1}\circ \nabla_\lambda H({\boldsymbol{\theta}},{\boldsymbol{\lambda}}),\\[0.3em]
Y({\boldsymbol{\theta}},{\boldsymbol{\lambda}})
= \nabla_y G(f({\boldsymbol{\theta}}),{\boldsymbol{\lambda}})^{-1}\circ \nabla_\lambda G(f({\boldsymbol{\theta}}),{\boldsymbol{\lambda}}).
\end{cases}
\end{align*}
We call $X$ the \textbf{characteristic direction} in parameter space and $Y$ the \textbf{characteristic output} in output space.
\end{definition}

Informally, $X({\boldsymbol{\theta}},{\boldsymbol{\lambda}})$ captures how an infinitesimal change in ${\boldsymbol{\lambda}}$ induces a direction in parameter space after ``solving for'' the corresponding change through $H$; $Y({\boldsymbol{\theta}},{\boldsymbol{\lambda}})$ is the analogous quantity on the output side through $G$. With these definitions in place, we state the main result for continuous equivariances.

\begin{theorem}[First- and second-order identities]\label{thm:main}
Assume $f$ admits a continuous $(H,G)$-equivariance and $({\boldsymbol{\theta}},{\boldsymbol{\lambda}})$ is a good position. Then the following hold:
\begin{enumerate}[leftmargin=2.0em, itemsep=0.4em]
    \item \textbf{(First-order identity).} The gradient of $L$ satisfies
    \begin{myeqwrapper}{i}\label{eq:main-first}
    \begin{flalign*}
    \text{(i).}\quad
    \nabla L({\boldsymbol{\theta}})\circ X({\boldsymbol{\theta}},{\boldsymbol{\lambda}})
    = \nabla \ell\big(f({\boldsymbol{\theta}})\big)\circ Y({\boldsymbol{\theta}},{\boldsymbol{\lambda}}). &&
    \end{flalign*}
    \end{myeqwrapper}

    \item \textbf{(Second-order identities).} The Hessian of $L$ satisfies the following two equations:
    \begin{myeqwrapper}{ii}\label{eq:main-second-1}
    \begin{flalign*}
    \text{(ii).}\quad
    \nabla^2 L({\boldsymbol{\theta}})\circ X({\boldsymbol{\theta}},{\boldsymbol{\lambda}})
    =&~
    \color{color2}\nabla^2 \ell\big(f({\boldsymbol{\theta}})\big)\circ Y({\boldsymbol{\theta}},{\boldsymbol{\lambda}})\circ_2 \nabla f({\boldsymbol{\theta}}) &&
    \\ &
    - \nabla \ell\big(f({\boldsymbol{\theta}})\big)\circ \nabla f({\boldsymbol{\theta}})\circ \nabla_\theta H({\boldsymbol{\theta}},{\boldsymbol{\lambda}})^{-1}\circ \nabla_\lambda \nabla_\theta H({\boldsymbol{\theta}},{\boldsymbol{\lambda}})
    \\ &
    \color{color1}+ \nabla \ell\big(f({\boldsymbol{\theta}})\big)\circ \nabla f({\boldsymbol{\theta}})\circ \nabla_\theta H({\boldsymbol{\theta}},{\boldsymbol{\lambda}})^{-1}\circ \nabla_\theta^2 H({\boldsymbol{\theta}},{\boldsymbol{\lambda}})\circ X({\boldsymbol{\theta}},{\boldsymbol{\lambda}})
    \\ & \color{color2}
    + \nabla \ell\big(f({\boldsymbol{\theta}})\big)\circ \nabla_y G(f({\boldsymbol{\theta}}),{\boldsymbol{\lambda}})^{-1}\circ \nabla_\lambda\nabla_y G(f({\boldsymbol{\theta}}),{\boldsymbol{\lambda}})\circ_2 \nabla f({\boldsymbol{\theta}})
    \\ & \color{color2}
    - \nabla \ell\big(f({\boldsymbol{\theta}})\big)\circ \nabla_y G(f({\boldsymbol{\theta}}),{\boldsymbol{\lambda}})^{-1}\circ \nabla_y^2 G(f({\boldsymbol{\theta}}),{\boldsymbol{\lambda}})\circ Y({\boldsymbol{\theta}},{\boldsymbol{\lambda}})\circ_2 \nabla f({\boldsymbol{\theta}}).
    \end{flalign*}
    \end{myeqwrapper}

    \begin{myeqwrapper}{iii}\label{eq:main-second-2}
    \begin{flalign*}
    \text{(iii).}\quad
    \nabla^2 L({\boldsymbol{\theta}})\circ X({\boldsymbol{\theta}},{\boldsymbol{\lambda}})\circ_2 X({\boldsymbol{\theta}},{\boldsymbol{\lambda}})
    =&\; \color{color2}
    \nabla^2 \ell\big(f({\boldsymbol{\theta}})\big)\circ Y({\boldsymbol{\theta}},{\boldsymbol{\lambda}})\circ_2 Y({\boldsymbol{\theta}},{\boldsymbol{\lambda}}) &&
    \\ &
    - \nabla \ell\big(f({\boldsymbol{\theta}})\big)\circ \nabla f({\boldsymbol{\theta}})\circ \nabla_\theta H({\boldsymbol{\theta}},{\boldsymbol{\lambda}})^{-1}\circ \nabla_\lambda^2 H({\boldsymbol{\theta}},{\boldsymbol{\lambda}})
    \\ & \color{color1}
    + \nabla \ell\big(f({\boldsymbol{\theta}})\big)\circ \nabla f({\boldsymbol{\theta}})\circ \nabla_\theta H({\boldsymbol{\theta}},{\boldsymbol{\lambda}})^{-1}\circ \nabla_\theta^2 H({\boldsymbol{\theta}},{\boldsymbol{\lambda}})\circ X({\boldsymbol{\theta}},{\boldsymbol{\lambda}})\circ_2 X({\boldsymbol{\theta}},{\boldsymbol{\lambda}})
    \\ & \color{color2}
    + \nabla \ell\big(f({\boldsymbol{\theta}})\big)\circ \nabla_y G(f({\boldsymbol{\theta}}),{\boldsymbol{\lambda}})^{-1}\circ \nabla_\lambda^2 G(f({\boldsymbol{\theta}}),{\boldsymbol{\lambda}})
    \\ & \color{color2}
    - \nabla \ell\big(f({\boldsymbol{\theta}})\big)\circ \nabla_y G(f({\boldsymbol{\theta}}),{\boldsymbol{\lambda}})^{-1}\circ \nabla_y^2 G(f({\boldsymbol{\theta}}),{\boldsymbol{\lambda}})\circ Y({\boldsymbol{\theta}},{\boldsymbol{\lambda}})\circ_2 Y({\boldsymbol{\theta}},{\boldsymbol{\lambda}}).
    \end{flalign*}
    \end{myeqwrapper}
\end{enumerate}
\end{theorem}

Although the formulas in \Cref{thm:main} appear complicated, they admit a simple interpretation and often simplify substantially. \Cref{eq:main-first} states that the directional derivative of $L$ along the characteristic direction $X$ is determined entirely by output-side quantities: the loss gradient $\nabla \ell(f({\boldsymbol{\theta}}))$ and the characteristic output $Y$, which are both zeroth-order properties of $f$. The second-order identities \cref{eq:main-second-1,eq:main-second-2} provide analogous constraints for the Hessian, expressing its action and quadratic form along $X$ in terms of lower-order information.

Two common special cases lead to significant simplification. First, when the equivariance is a \emph{symmetry}, we have $\nabla_\lambda G \equiv 0$ and hence $Y \equiv 0$; all terms shown in {\color{color2}green} vanish. Second, when $H({\boldsymbol{\theta}},{\boldsymbol{\lambda}})$ is \emph{linear} in ${\boldsymbol{\theta}}$, we have $\nabla_\theta^2 H = 0$; all terms shown in {\color{color1}blue} vanish. Many equivariances of practical interest satisfy one or both conditions, yielding compact and interpretable identities.

\subsection{Discrete Symmetry}

When the transformation $H$ is not continuously parameterized in ${\boldsymbol{\lambda}}$ (for instance, when ${\boldsymbol{\lambda}}$ indexes a discrete set of transformations such as sign flips or permutations), the continuous identities of \Cref{thm:main} do not directly apply. Nevertheless, the equivariance condition still implies useful constraints when we restrict attention to the fixed-point set of a given transformation. This approach is common in the study of discrete symmetries \citep{brea_weight_space_permuation,ziyin_mirror_symmetry}. We first provide a formal definition of the fixed-point set.

\begin{definition}[Fixed-point set]\label{def:fixed-point}
For ${\boldsymbol{\lambda}}_0 \in \mathbb R^p$, we say that $\mathcal S \subseteq \mathbb R^d$ is a \textbf{fixed-point set} of $H$ at ${\boldsymbol{\lambda}}_0$ if $H({\boldsymbol{\theta}}, {\boldsymbol{\lambda}}_0) = {\boldsymbol{\theta}}$ for all ${\boldsymbol{\theta}} \in \mathcal S$. We denote the union of all fixed-point sets of $H$ at ${\boldsymbol{\lambda}}_0$ by $\mathrm{Fix}(H, {\boldsymbol{\lambda}}_0)$.
\end{definition}

In this subsection, we focus on \emph{symmetry}, i.e., $G = \mathrm{Id}$. In most cases of discrete symmetry, $H(\cdot, {\boldsymbol{\lambda}}_0)$ is a linear transformation which only has eigenvalues $\pm 1$ (e.g. permutations); the eigenspace corresponding to eigenvalue $1$ is then a fixed-point set. Below we present our results for discrete symmetries.

\begin{theorem}[Discrete symmetry identities]\label{thm:main-discrete}
Suppose $f$ admits an $(H,\mathrm{Id})$-equivariance, i.e., $f(H({\boldsymbol{\theta}},{\boldsymbol{\lambda}})) = f({\boldsymbol{\theta}})$ for all ${\boldsymbol{\theta}}$ and ${\boldsymbol{\lambda}}$. Fix ${\boldsymbol{\lambda}}_0 \in \mathbb R^p$ and let ${\boldsymbol{\theta}} \in \mathrm{Fix}(H, {\boldsymbol{\lambda}}_0)$. Define $\widehat X({\boldsymbol{\theta}}, {\boldsymbol{\lambda}}_0) := \nabla_{\boldsymbol{\theta}} H({\boldsymbol{\theta}}, {\boldsymbol{\lambda}}_0)$. Then:
\begin{enumerate}[leftmargin=2.0em, itemsep=0.4em]
    \item \textbf{(First-order identity).} The gradient of $L$ satisfies
    \begin{myeqwrapper}{i'}\label{eq:discrete-first}
    \begin{flalign*}
    \text{(i').}\quad
    \nabla L({\boldsymbol{\theta}})\circ \widehat X({\boldsymbol{\theta}},{\boldsymbol{\lambda}}_0)
    = \nabla L({\boldsymbol{\theta}}). &&
    \end{flalign*}
    \end{myeqwrapper}

    \item \textbf{(Second-order identity).} The Hessian of $L$ satisfies:
    \begin{myeqwrapper}{ii'}\label{eq:discrete-second}
    \begin{flalign*}
    \text{(ii').}\quad
    \nabla^2 L({\boldsymbol{\theta}})\circ \widehat X({\boldsymbol{\theta}},{\boldsymbol{\lambda}}_0)\circ_2 \widehat X({\boldsymbol{\theta}},{\boldsymbol{\lambda}}_0)
    = \nabla^2 L({\boldsymbol{\theta}}) \color{color1}{ - \nabla L({\boldsymbol{\theta}}) \circ \nabla_{\theta}^2 H({\boldsymbol{\theta}}, {\boldsymbol{\lambda}}_0)}. && 
    \end{flalign*}
    \end{myeqwrapper}
\end{enumerate}
\end{theorem}

The key difference from the continuous case is that here $\widehat X = \nabla_{\boldsymbol{\theta}} H$ is the Jacobian of $H$ with respect to ${\boldsymbol{\theta}}$ (not an inverse times a ${\boldsymbol{\lambda}}$-derivative), and the identities hold only on the fixed-point set where $H({\boldsymbol{\theta}}, {\boldsymbol{\lambda}}_0) = {\boldsymbol{\theta}}$.

\Cref{eq:discrete-first} states that the gradient is an eigenvector of $\widehat X^\top$ with eigenvalue $1$: the transformation $H$ leaves the gradient invariant (when viewed as a linear functional). \Cref{eq:discrete-second} constrains the structure of the Hessian: conjugating by $\widehat X$ reproduces the original Hessian up to a correction term involving $\nabla L$ and the second derivative $\nabla_{\boldsymbol{\theta}}^2 H$. When $H$ is linear in ${\boldsymbol{\theta}}$ (so that $\nabla_{\boldsymbol{\theta}}^2 H = 0$), the {\color{color1}blue} term vanishes, and the identity simplifies to $\widehat X^\top \nabla^2 L \, \widehat X = \nabla^2 L$, i.e., the Hessian is invariant under conjugation by the transformation Jacobian.

These discrete identities are weaker than their continuous counterparts (they hold only on the fixed-point set rather than everywhere), but they still yield nontrivial structural information. In \Cref{sec:discrete}, we show that they recover recent results on mirror symmetry \citep{ziyin_mirror_symmetry} as a special case.

\section{First-Order Consequences: Conservation Laws and Implicit Bias}\label{sec:firstorder}

We now instantiate the first-order identity \cref{eq:main-first} in concrete settings. This section shows that many known results in learning dynamics, including conservation laws \citep{bozhao_conserved,noether}, gradient flow invariants \citep{weihu_deep_linear,relu_balance,relu_balance_2}, and scalar identities for implicit bias \citep{kaifeng_homo}, arise as special cases of \cref{eq:main-first}.

\subsection{Symmetry}\label{sec:symmetry-first-order}

When $G({\boldsymbol{y}},{\boldsymbol{\lambda}})={\boldsymbol{y}}$ (symmetry), we have $Y \equiv {\boldsymbol{0}}$, and \cref{eq:main-first} reduces to
\begin{align}
\left\langle \nabla L({\boldsymbol{\theta}}),\, X({\boldsymbol{\theta}},{\boldsymbol{\lambda}})\right\rangle = 0.
\label{eq:first-sym-ortho}
\end{align}
If the model is trained through gradient flow (GF), we have $\dot{\boldsymbol{\theta}}=-\nabla L({\boldsymbol{\theta}})$, and the identity becomes $\langle \dot{\boldsymbol{\theta}}, X({\boldsymbol{\theta}},{\boldsymbol{\lambda}})\rangle = 0$: the parameter motion is orthogonal to the symmetry-generated direction. Integrating this orthogonality along the flow yields conservation laws, recovering Noether-type results \citep{noether,hidenori_symmetry}. Two representative examples are:\footnote{For notational simplicity, when focusing on a subset of parameters, we view $f$ as a function of those parameters.}
\begin{enumerate}[leftmargin=1.5em, itemsep=0.3em]
    \item \textbf{Rescaling symmetry.} If two consecutive layers with trainable parameters ${\boldsymbol{W}}_1,{\boldsymbol{W}}_2$ satisfy $f({\boldsymbol{W}}_1,{\boldsymbol{W}}_2)=f(\lambda{\boldsymbol{W}}_1,\lambda^{-1}{\boldsymbol{W}}_2)$, then \cref{eq:first-sym-ortho} implies $\|{\boldsymbol{W}}_1\|_F^2-\|{\boldsymbol{W}}_2\|_F^2$ is constant during training \citep{andrew_saxe_exact}. This conservation law is often used in analyses of two-layer ReLU networks to tie the absolute value of the second-layer coefficients directly with the norm of the first-layer parameters and eliminates its freedom \citep{relu_balance,relu_balance_2}.

    \item \textbf{Reparameterization symmetry.} If a network contains a component $f_0({\boldsymbol{W}}_1, {\boldsymbol{W}}_2) = {\boldsymbol{W}}_1 {\boldsymbol{W}}_2$, then for any invertible matrix ${\boldsymbol{P}}$, we have $f({\boldsymbol{P}}{\boldsymbol{W}}_1, {\boldsymbol{P}}^{-1}{\boldsymbol{W}}_2) = f({\boldsymbol{W}}_1, {\boldsymbol{W}}_2)$. This reparameterization symmetry appears in deep linear networks and self-attention mechanisms. \Cref{eq:first-sym-ortho} then implies ${\boldsymbol{W}}_1 {\boldsymbol{W}}_1^\top - {\boldsymbol{W}}_2 {\boldsymbol{W}}_2^\top = \mathrm{const}$, which has been used to analyze solution stability \citep{deep-linaer-sharpness,weihu_deep_linear,vidal_deep_linear,wei_auto_balanced,andrew_saxe_balanced}.
\end{enumerate}

In discrete-time gradient descent, ${\boldsymbol{\theta}}_{t+1}={\boldsymbol{\theta}}_t-\eta \nabla L({\boldsymbol{\theta}}_t)$, the identity \cref{eq:first-sym-ortho} still implies that the update is orthogonal to the characteristic direction at each iterate, capturing the same geometric constraint on the update direction.

\subsection{Homogeneity}\label{sec:first-order-homo}

We say $f$ is \textbf{$m$-homogeneous} if $f(\lambda {\boldsymbol{\theta}})=\lambda^m f({\boldsymbol{\theta}})$ for all $\lambda\in \mathbb R$. For example, an $m$-layer ReLU network without bias is $m$-homogeneous. Homogeneity fits our framework with $H({\boldsymbol{\theta}},\lambda)=\lambda {\boldsymbol{\theta}}$ and $G({\boldsymbol{y}},\lambda)=\lambda^m {\boldsymbol{y}}$. At any good position with $\lambda\neq 0$, we have $X({\boldsymbol{\theta}},\lambda)=\lambda^{-1}{\boldsymbol{\theta}}$ and $Y({\boldsymbol{\theta}},\lambda)= m\lambda^{-1} f({\boldsymbol{\theta}})$. \Cref{eq:main-first} then yields the well-known Euler-type relation
\begin{align*}
\left\langle \nabla L({\boldsymbol{\theta}}),\,{\boldsymbol{\theta}}\right\rangle
= m \left\langle \nabla \ell\big(f({\boldsymbol{\theta}})\big),\, f({\boldsymbol{\theta}})\right\rangle.
\end{align*}
Under gradient flow and specific loss functions, this identity translates into a monotonicity statement. To see this, let $c = 1$ and $\dot{\boldsymbol{\theta}}=-\nabla L({\boldsymbol{\theta}})$, we have
\begin{align*}
\frac{\mathrm d}{\mathrm dt}\frac{1}{2}\|{\boldsymbol{\theta}}\|^2
=-\left\langle {\boldsymbol{\theta}},\nabla L({\boldsymbol{\theta}})\right\rangle
=-m\,\ell'\big(f({\boldsymbol{\theta}})\big)\, f({\boldsymbol{\theta}}).
\end{align*}
For classification losses where $\ell'(y) \cdot y < 0$ when correctly classified (e.g., exponential or logistic loss), once the model enters the correct-classification regime, the right-hand side is positive, so $\|{\boldsymbol{\theta}}(t)\|$ diverges. This norm divergence is the key step in max-margin implicit bias proofs \citep{kaifeng_homo}.

\paragraph{Beyond these two examples.} The two cases above already capture the general workflow of the toolbox. Once a continuous equivariance is specified, \cref{eq:main-first} immediately yields a scalar identity that can be turned into either 1) an invariant along gradient flow (in the symmetry case) or 2) a differential inequality controlling the evolution of a meaningful quantity such as the parameter norm or margin (in the non-symmetric case). Such identities are exactly the starting point of many implicit-bias and stability arguments for gradient methods. We emphasize again that the framework is not restricted to $G=\mathrm{Id}$: allowing general equivariances produces non-vanishing right-hand sides in \cref{eq:main-first}.

\section{Second-Order Consequences: Hessian Structure and Loss Landscape}\label{sec:secondorder}

While the first-order results in \Cref{sec:firstorder} unify many known phenomena, most of those consequences have been established (albeit separately) in the literature. The second-order implications of equivariance, however, have not been explored nearly as systematically, which can be essential for studying the geometry of optimization in modern networks, as they translate symmetry/equivariance directly into testable structural predictions about curvature.

In this section, we focus on the second-order identities \cref{eq:main-second-1,eq:main-second-2} that constrain the Hessian $\nabla^2 L({\boldsymbol{\theta}})$ through its action and quadratic forms along the same transformation-induced directions. This is precisely the level at which many empirical observations are formulated: sharp versus flat directions, low-rank or low-dimensional structure of curvature, and the alignment between optimization dynamics and the leading eigenspaces of the Hessian \citep{gd_tiny_space,eigval_of_hessian,sharpness_dynamics,early-stage-sharpness}. Below we illustrate the powerfulness of the second-order identities through several examples.

\subsection{Homogeneity}\label{sec:secondorder-homo}

For $m$-homogeneous models (discussed in \Cref{sec:first-order-homo}), recall that $X({\boldsymbol{\theta}},\lambda)=\lambda^{-1}{\boldsymbol{\theta}}$ and $Y({\boldsymbol{\theta}},\lambda)=m\lambda^{-1}y$ where $y=f({\boldsymbol{\theta}})$. As in \Cref{sec:first-order-homo}, here we consider the case where $c=1$ for simplicity of the illustration. Substituting the equivariance into \cref{eq:main-second-1,eq:main-second-2} yields the following identities:
\begin{align}
\nabla^2 L({\boldsymbol{\theta}})\,{\boldsymbol{\theta}}
&=
\left(m y\frac{\ell''(y)}{\ell'(y)} + m-1\right)\nabla L({\boldsymbol{\theta}}),
\label{eq:realization-second-homo-1}
\\
\left\langle {\boldsymbol{\theta}},\,\nabla^2 L({\boldsymbol{\theta}})\,{\boldsymbol{\theta}}\right\rangle
&=
\ell''(y)\,m^2 y^2+\ell'(y)\,m(m-1)\,y,
\label{eq:realization-second-homo-2}
\end{align}
whenever $\ell'(y)\neq 0$. These identities themselves already provide strong information about the structure of the Hessian and the Hessian-gradient relationship, as they relate the action (and quadratic form) of the Hessian along ${\boldsymbol{\theta}}$ to low-order quantities.

\paragraph{Gradient descent happens in a tiny subspace.}
By performing eigen-decomposition of the Hessian, the first identity \cref{eq:realization-second-homo-1} immediately gives the following corollary.

\begin{corollary}\label{cor:gradient-hessian-eigen-decomp}
    For a $m$-homogeneous $f$ and a good position $({\boldsymbol{\theta}}, {\boldsymbol{\lambda}})$, there exists $\alpha({\boldsymbol{\theta}}) \in \mathbb R$ such that
    $\langle{\boldsymbol{g}}, {\boldsymbol{u}}_k\rangle = \lambda_k \alpha({\boldsymbol{\theta}})  \langle{\boldsymbol{\theta}}, {\boldsymbol{u}}_k\rangle$,
    where ${\boldsymbol{g}} = \nabla L({\boldsymbol{\theta}})$ and $(\lambda_k,{\boldsymbol{u}}_k)$ are eigenvalue-eigenvector pairs of $\nabla^2 L({\boldsymbol{\theta}})$.
\end{corollary}

\Cref{cor:gradient-hessian-eigen-decomp} reveals a deep connection of the angles between the gradient, eigenvector of the Hessian, and the parameter. A first consequence is that $\nabla L({\boldsymbol{\theta}})$ must lie in the column space of $\nabla^2 L({\boldsymbol{\theta}})$, which has been proven to be low-dimensional around the stable points \citep{hessian_structure,hessian_singular}. Moreover, in regimes where ${\boldsymbol{\theta}}$ does not concentrate on a particular subset of eigenvectors (e.g., as a rough heuristic, $\langle{\boldsymbol{\theta}},{\boldsymbol{u}}_k\rangle$ being of comparable magnitude across many $k$), the gradient preferentially aligns with directions of large $|\lambda_k|$, concentrating in leading eigenspaces. This provides a concrete structural explanation for the empirical observation that gradient descent effectively evolves within a small subspace associated with dominant curvature directions \citep{gd_tiny_space}.

\paragraph{Sharpness lower bound.}
The second identity \cref{eq:realization-second-homo-2} yields an immediate curvature lower bound along the direction of ${\boldsymbol{\theta}}$. Let $K$ denote the sharpness of the loss (i.e. largest eigenvalue of $\nabla^2 L({\boldsymbol{\theta}})$), \cref{eq:realization-second-homo-2} gives a computable lower bound on $K$:
\begin{align*}
K=\sup_{{\boldsymbol{x}}\neq{\boldsymbol{0}}}\frac{\left\langle{\boldsymbol{x}},\nabla^2 L({\boldsymbol{\theta}}){\boldsymbol{x}}\right\rangle}{\|{\boldsymbol{x}}\|^2} 
\  \ge\
\frac{\left\langle{\boldsymbol{\theta}},\nabla^2 L({\boldsymbol{\theta}}){\boldsymbol{\theta}}\right\rangle}{\|{\boldsymbol{\theta}}\|^2}
 = \frac{m}{\|{\boldsymbol{\theta}}\|^2}\Big(\ell''(y)\,m y^2+\ell'(y)\,(m-1)\,y\Big).
\end{align*}
Although this bound probes only a single direction, it is a particularly canonical one for homogeneous models: scaling ${\boldsymbol{\theta}}$ changes the output in a controlled manner, and thus curvature along ${\boldsymbol{\theta}}$ captures a fundamental component of the loss geometry induced by homogeneity.

As an example of a concrete setting, consider $c=1, m=1$ and the exponential loss $\ell(y)=\exp(-y \hat y)$ for binary classification, this becomes $K \geq \frac{\ell(y)y^2}{\|{\boldsymbol{\theta}}\|^2}$. In early training when $|y|$ is small, the lower bound is dominated by the $y^2$ term, suggesting a monotonic growth of the sharpness lower bound as training goes, partially verifying the previously observed progressive sharpening phenomenon \citep{eos}.

\subsection{Noether Flow under Stochastic Dynamics}\label{sec:noether-flow}

Now, we consider the setting where the equivariance is a symmetry ($G = \mathrm{Id}$) and the training is stochastic gradient flow (SGF).  Recall from \Cref{sec:symmetry-first-order} that under gradient flow, symmetries induce conserved quantities. Specifically, assume $p=1$ and $H({\boldsymbol{\theta}},0) = {\boldsymbol{\theta}}$ (so $\lambda=0$ corresponds to the identity transformation). Then $X({\boldsymbol{\theta}},0) = \nabla_{\lambda} H({\boldsymbol{\theta}},0)$. If there exists $C: \mathbb R^d \to \mathbb R$ with
\begin{align}
\nabla C({\boldsymbol{\theta}}) = \nabla_\lambda H({\boldsymbol{\theta}},0), \label{eq:def-consevred-quantity}
\end{align}
then $\dot C({\boldsymbol{\theta}}) = \langle \nabla C({\boldsymbol{\theta}}), \dot{\boldsymbol{\theta}} \rangle = -\langle \nabla_{\lambda} H({\boldsymbol{\theta}},0), \nabla L({\boldsymbol{\theta}}) \rangle = 0$, which means $C$ is conserved under GF (called a Noether charge).

Under SGF, however, $C$ is no longer conserved but can exhibit a systematic drift called \emph{Noether flow} \citep{ziyin_sgd}. Our framework provides a precise characterization of this drift. Specifically, in this subsection we follow the setting of \citep{ziyin_sgd}, which adopts a multi-sample setting. Let $\Omega$ be the set of all possible input data and $\mu$ be a distribution over $\Omega$. For a data sample $x \in \Omega$, let $L_x({\boldsymbol{\theta}})$ represent the loss function with data $x$ and trainable parameter ${\boldsymbol{\theta}}$. Let $L({\boldsymbol{\theta}}) = \mathbb E_{x \sim \mu} L_x({\boldsymbol{\theta}})$ be the expected loss function. Let $\Sigma({\boldsymbol{\theta}}) = \mathbb E\left(\nabla L_x({\boldsymbol{\theta}}) \nabla L_x({\boldsymbol{\theta}})^\top\right) - \nabla  L({\boldsymbol{\theta}})\nabla  L({\boldsymbol{\theta}})^\top$ be the noise covariance matrix. Specifically, we can obtain the following characterization of the Noether flow using \Cref{thm:main}:

\begin{corollary}\label{cor:neother-flow}
    If $L$ is sufficiently smooth, and the parameter ${\boldsymbol{\theta}}$ is trained through SGF (defined in \citep{ziyin_sgd}), the conserved quantity $C$ from \cref{eq:def-consevred-quantity} evolves as
    \begin{align*}
    \dot C({\boldsymbol{\theta}}) = - \frac{\sigma^2}{2} \left\langle\nabla C({\boldsymbol{\theta}}), \frac{\partial }{\partial {\boldsymbol{\theta}}} \mathrm{Tr}(\Sigma({\boldsymbol{\theta}}))\right\rangle,
    \end{align*}
    where $\Sigma({\boldsymbol{\theta}})$ is the noise covariance matrix and $\sigma$ is the strength of the noise in the SGF.
\end{corollary}

\Cref{cor:neother-flow} shows that the Noether charge drifts in a direction determined by how the total gradient noise $\mathrm{Tr}(\Sigma({\boldsymbol{\theta}}))$ changes along the symmetry direction $\nabla C({\boldsymbol{\theta}})$. If noise increases along $\nabla C$, the charge decreases, and vice versa. This provides a precise mechanism for how stochasticity breaks conservation laws: the drift is not arbitrary but is governed by the interaction between the symmetry structure and the noise geometry.

\subsection{Last-Layer Equivariance}\label{sec:secondorder-lastlayer}

We now consider an equivariance that is neither a symmetry nor homogeneity. Consider a model of the form
\begin{align}
f({\boldsymbol{W}},{\boldsymbol{\theta}}')= {\boldsymbol{W}}h({\boldsymbol{\theta}}'),\label{eq:last-layer}
\end{align}
where ${\boldsymbol{W}}\in \mathbb R^{c \times s}$ is the last-layer weight matrix, ${\boldsymbol{\theta}}' \in \mathbb R^{d'}$ contains the remaining parameters, and $h: \mathbb R^{d'} \to \mathbb R^s$ represents the feature extractor, $d' = d-cs$ is the number of parameters of the model excluding the last layer. For any ${\boldsymbol{P}} \in \mathbb R^{c \times c}$, the model admits the equivariance $f({\boldsymbol{P}}{\boldsymbol{W}}, {\boldsymbol{\theta}}') = {\boldsymbol{P}} f({\boldsymbol{W}}, {\boldsymbol{\theta}}')$, fitting our framework with $H(({\boldsymbol{W}},{\boldsymbol{\theta}}'), {\boldsymbol{P}}) = ({\boldsymbol{P}}{\boldsymbol{W}}, {\boldsymbol{\theta}}')$ and $G({\boldsymbol{y}} {\boldsymbol{P}}) = {\boldsymbol{P}}{\boldsymbol{y}}$.

Applying this equivariance to the the second-order identity \cref{eq:main-second-2} yields the following result:
\begin{corollary}\label{cor:last-layer-alignment}
    For $f$ satisfying \cref{eq:last-layer} and any ${\boldsymbol{V}} \in \mathbb R^{c\times s}$ whose row space is contained in that of ${\boldsymbol{W}}$,
    \begin{align}
    \langle\mathop{ \mathrm{ vec } } {\boldsymbol{V}}, \nabla^2_{\mathop{ \mathrm{ vec } } {\boldsymbol{W}}} L({\boldsymbol{\theta}}) \mathop{ \mathrm{ vec } } {\boldsymbol{V}}\rangle = \langle {\boldsymbol{V}}{\boldsymbol{h}}, \nabla^2\ell({\boldsymbol{y}}) {\boldsymbol{V}}{\boldsymbol{h}} \rangle, \label{eq:last-layer-alignment-eq}
    \end{align}
    where ${\boldsymbol{h}} = h({\boldsymbol{\theta}}')$, ${\boldsymbol{y}} = f({\boldsymbol{\theta}})$, and ${\boldsymbol{\theta}} = \left({\boldsymbol{W}}, {\boldsymbol{\theta}}'\right)$.
\end{corollary}

This identity directly connect the parameter space sharpness along direction ${\boldsymbol{V}}$ with the output space sharpness along direction ${\boldsymbol{V}}{\boldsymbol{h}}$. This connection could be potentially related to the self-duality structure of the last layer weight matrix observed in \citep{nc}.

As a specific example, consider the softmax cross-entropy loss. The Hessian of $\ell$ satisfies $\nabla^2\ell({\boldsymbol{y}})=\mathrm{diag}({\boldsymbol{p}})-{\boldsymbol{p}}{\boldsymbol{p}}^\top$ where ${\boldsymbol{p}}=\mathrm{softmax}({\boldsymbol{y}})$, so \cref{eq:last-layer-alignment-eq} becomes
$\langle\mathop{ \mathrm{ vec } } {\boldsymbol{V}}, \nabla^2_{\mathop{ \mathrm{ vec } } {\boldsymbol{W}}} L \, \mathop{ \mathrm{ vec } } {\boldsymbol{V}}\rangle = \mathrm{Var}_{{\boldsymbol{p}}}({\boldsymbol{V}}{\boldsymbol{h}})$,
where $\mathrm{Var}_{{\boldsymbol{p}}}$ is the variance under the distribution ${\boldsymbol{p}}$. As training progresses and ${\boldsymbol{p}}$ concentrates on the true class $k$, perturbations that redistribute mass among low-probability classes have vanishing curvature (flat directions), while perturbations affecting the top-vs-rest margin have large curvature (sharp directions). This explains why late-stage last-layer geometry becomes increasingly rigid, with directions stabilized up to rescaling \citep{nc,nc-ufm}.

\subsection{Symmetry: Hessian Degeneracy at Stationarity}\label{sec:secondorder-symmetry}

In \Cref{sec:noether-flow} we examined how symmetries affect dynamics under stochastic training. Here we consider a complementary aspect: what symmetries imply about the Hessian at stationary points.

For symmetry ($G = \mathrm{Id}$), the characteristic output $Y \equiv {\boldsymbol{0}}$. At any first-order stable point ${\boldsymbol{\theta}}_\star$ (where $\nabla L({\boldsymbol{\theta}}_\star)={\boldsymbol{0}}$) we have
\begin{align*}
\nabla^2 L({\boldsymbol{\theta}}_\star) \circ X({\boldsymbol{\theta}}_\star,{\boldsymbol{\lambda}})={\boldsymbol{0}}.
\end{align*}
Thus, every symmetry-generated direction $X({\boldsymbol{\theta}}_\star,{\boldsymbol{\lambda}})$ lies in the null space of the Hessian at stationarity. This is consistent with the principle that continuous symmetries create degenerate manifolds of equivalent solutions \citep{hessian-illness}. Consequently, if a model has $q$ independent continuous symmetries whose characteristic directions span a $q$-dimensional subspace at ${\boldsymbol{\theta}}_\star$, then $\nabla^2 L({\boldsymbol{\theta}}_\star)$ has at least $p$ zero eigenvalues. This provides a structural explanation for why Hessians of over-parameterized networks often exhibit large near-zero spectral bulk \citep{hessian_structure,hessian-lowrank,empirical-hessian-lowrank,empirical-hessian-lowrank-2,hessian-lowrank-2}.

We note that previous work \citep{hidenori_symmetry} has derived equivalent second-order identities for continuous symmetries but without much discussion; our contribution here is then to place these results within the broader equivariance framework that also handles non-symmetric and non-continuous cases.

\section{Discrete Symmetry Consequences}\label{sec:discrete}

The preceding sections focused on smooth equivariances, where the transformation parameter ${\boldsymbol{\lambda}}$ varies continuously. However, many symmetries in neural networks are inherently discrete: sign flips, permutations of hidden neurons, or reflections across subspaces. For such transformations, the continuous identities of \Cref{thm:main} do not directly apply since we cannot differentiate with respect to ${\boldsymbol{\lambda}}$. Nevertheless, \Cref{thm:main-discrete} shows that useful constraints still hold on the fixed-point set of the transformation.

Specifically, consider a discrete symmetry $H$ and a point ${\boldsymbol{\lambda}}_0$, assume $H(\cdot, {\boldsymbol{\lambda}}_0)$ is a linear transformation of the parameters, i.e.  $H({\boldsymbol{\theta}}, {\boldsymbol{\lambda}}_0) = {\boldsymbol{P}} {\boldsymbol{\theta}}$  for some fixed matrix ${\boldsymbol{P}}$ (depending on ${\boldsymbol{\lambda}}_0$). Since $G = \mathrm{Id}$ for symmetry and $H$ is linear in ${\boldsymbol{\theta}}$, \Cref{thm:main-discrete} simplifies to
\begin{align}
{\boldsymbol{P}}^\top \nabla L({\boldsymbol{\theta}}) & = \nabla L({\boldsymbol{\theta}}), \label{eq:dis-realization-first}
\\ {\boldsymbol{P}}^\top \nabla^2 L({\boldsymbol{\theta}}) \, {\boldsymbol{P}} & = \nabla^2 L({\boldsymbol{\theta}}), \label{eq:dis-realization-second}
\end{align}
for any ${\boldsymbol{\theta}} \in \mathrm{Fix}(H, {\boldsymbol{\lambda}}_0)$.

These identities have clear interpretations. \Cref{eq:dis-realization-first} states that the gradient is an eigenvector of ${\boldsymbol{P}}^\top$ with eigenvalue $1$: the symmetry leaves the gradient invariant on the fixed-point set. \Cref{eq:dis-realization-second} states that the Hessian commutes with ${\boldsymbol{P}}$ (in the sense of conjugation): the curvature respects the same symmetry structure as the transformation.

An important consequence is that eigenvectors of ${\boldsymbol{P}}$ organize the Hessian's eigenspaces. If ${\boldsymbol{P}}$ has distinct eigenvalues, then eigenvectors of $\nabla^2 L({\boldsymbol{\theta}})$ must align with eigenvectors of ${\boldsymbol{P}}$. More generally, \cref{eq:dis-realization-second} implies that the Hessian block-diagonalizes according to the eigenspaces of ${\boldsymbol{P}}$: directions corresponding to different eigenvalues of ${\boldsymbol{P}}$ are decoupled in the Hessian.

\subsection{Mirror Symmetry}

As a concrete application, we recover the mirror symmetry results of \citet{ziyin_mirror_symmetry} as a special case. Let ${\boldsymbol{O}} \in \mathbb R^{d \times d'}$ be a matrix with orthonormal columns ($d' \leq d$), defining a $d'$-dimensional subspace. The mirror transformation defined by ${\boldsymbol{O}}$ is ${\boldsymbol{P}} = {\boldsymbol{I}} - 2{\boldsymbol{O}}{\boldsymbol{O}}^\top$, which reflects vectors across the orthogonal complement of $\mathop{ \mathrm{ col } }({\boldsymbol{O}})$. The fixed-point set consists of all ${\boldsymbol{\theta}} \perp \mathop{ \mathrm{ col } }({\boldsymbol{O}})$. Applying \cref{eq:dis-realization-first,eq:dis-realization-second} yields the following result:

\begin{corollary}\label{cor:mirro-sym}
    If $f$ has $(H, \mathrm{Id})$-equivariance with $H({\boldsymbol{\theta}},{\boldsymbol{\lambda}}_0) = ({\boldsymbol{I}} - 2{\boldsymbol{O}}{\boldsymbol{O}}^\top){\boldsymbol{\theta}}$, then for any ${\boldsymbol{\theta}} \perp \mathop{ \mathrm{ col } }({\boldsymbol{O}})$:
    \begin{enumerate}[leftmargin=1.5em, itemsep=0.2em]
        \item ${\boldsymbol{O}}^\top \nabla L({\boldsymbol{\theta}}) = {\boldsymbol{0}}$, i.e., the gradient has no component in $\mathop{ \mathrm{ col } }({\boldsymbol{O}})$;
        \item The Hessian preserves the subspace decomposition: $\nabla^2 L({\boldsymbol{\theta}}) \, \mathop{ \mathrm{ col } }({\boldsymbol{O}}) \subseteq \mathop{ \mathrm{ col } }({\boldsymbol{O}})$ and $\nabla^2 L({\boldsymbol{\theta}}) \, \mathop{ \mathrm{ col } }({\boldsymbol{O}})^\perp \subseteq \mathop{ \mathrm{ col } }({\boldsymbol{O}})^\perp$.
    \end{enumerate}
\end{corollary}

\Cref{cor:mirro-sym} recovers Theorems 1.1 and 1.2 of \citet{ziyin_mirror_symmetry}, which is the key theoretical result of it.

\section{Discussion}\label{sec:discussion}
\vspace{-1em}

In this work, we have developed a general equivariance toolbox that provides coupled first- and second-order constraints on learning dynamics. The framework substantially extends classical Noether-type analyses in three directions: 1) from first-order to second-order, capturing Hessian structure beyond gradient orthogonality; 2) from symmetry to general equivariance, handling cases like homogeneity where outputs transform non-trivially; and 3) from continuous to discrete transformations, where fixed-point constraints replace differential identities. Together, these extensions yield a unified calculus for translating transformation structure into predictions about optimization geometry.

\vspace{-1em}
\paragraph{Beyond Noether: from symmetry to equivariance.}
Classical applications of Noether's theorem to learning dynamics focus on symmetries ($G = \mathrm{Id}$), deriving conservation laws from the orthogonality $\langle \nabla L, X \rangle = 0$. Our framework generalizes this by allowing non-trivial output transformations $G$, which produces the richer identity $\langle \nabla L, X \rangle = \langle \nabla \ell, Y \rangle$. This extension is essential for capturing equivariances like homogeneity, where the right-hand side is non-zero and drives the implicit bias. In this sense, the toolbox unifies two previously separate classes of results: Noether-type conservation laws and homogeneity-based scalar identities now emerge as special cases of the same first-order identity.

\vspace{-1em}
\paragraph{Second-order structure and curvature phenomena.}
A central contribution is that the same equivariance calculus yields second-order constraints (\Cref{thm:main}, \cref{eq:main-second-1,eq:main-second-2}). These provide structural information about $\nabla^2 L({\boldsymbol{\theta}})$: how the Hessian acts along characteristic directions and what its quadratic forms evaluate to. When $H$ and $G$ are linear in their main arguments, the general formulas simplify to compact expressions directly comparable with empirical observations. This connects transformation structure to modern curvature phenomena such as low-rank Hessians, progressive sharpening, and gradient concentration in leading eigenspaces.

\vspace{-1em}
\paragraph{Discrete symmetries and fixed-point constraints.}
The extension to discrete symmetries (\Cref{thm:main-discrete}) shows that useful constraints persist even without continuous parameterization. While continuous symmetries yield differential identities holding everywhere, discrete symmetries yield algebraic constraints on the fixed-point set: the gradient must be invariant under the transformation, and the Hessian must commute with it. These constraints predict block-diagonal Hessian structure aligned with symmetry eigenspaces, complementing the zero-eigenvalue predictions from continuous symmetries. The mirror symmetry results of \citet{ziyin_mirror_symmetry} emerge as a special case, illustrating how the framework recovers and contextualizes recent findings.

\vspace{-1em}
\paragraph{Limitations and future directions.}
While the equivariance identities in \Cref{thm:main,thm:main-discrete} are universal, turning them into sharp predictions in concrete learning problems still requires additional modeling choices. We highlight two directions that we view as particularly important.

\begin{itemize}[leftmargin=1.2em, itemsep=0.2em]

\vspace{-0.5em}
    \item \textbf{Training-dependent refinements.}
    Many of our current Hessian characterizations  are obtained by directly ``reading off'' structure from the identities, which is broadly applicable but largely agnostic to the training setting. In practice, data distributions, optimizer details (e.g., learning rate, momentum, weight decay, sharpness-aware updates), and architectural choices may determine which equivariances are effectively expressed and how strongly the induced constraints shape the trajectory. An important next step is to combine our identities with concrete training models to derive more quantitative predictions, such as evolution laws or scaling behaviors for curvature/sharpness along characteristic directions.

\vspace{-0.5em}
    \item \textbf{Beyond global equivariances: local and emergent structure.}
    Current analysis mainly emphasizes global equivariances/symmetries, whereas realistic training may exhibit additional local or approximate equivariances only in certain regimes (e.g., near particular solutions due to saturation, feature collapse, or local linearization). Since our framework is fundamentally local (as it only relies on derivatives around a good position), it can in principle accommodate such cases, but we do not yet provide a systematic characterization of when these emergent equivariances arise or how to validate them empirically. Developing practical criteria and second-order signatures for local equivariances may help further clarify end-stage phenomena such as progressive flattening and structured degeneracies.
\end{itemize}

Overall, the equivariance toolbox provides a unified interface between transformation structure and learning dynamics, extending classical Noether-type analyses to second-order constraints and discrete symmetries, and offering a systematic route to interpret modern empirical observations about optimization geometry.

\newpage
\bibliography{ref}
\bibliographystyle{plainnat}

\newpage
\appendix

\input{appendix}

\end{document}

%% file: appendix.tex
\section{Proofs of the theoretical results}

We first state some basic tensor-function operations to simplify the subsequent derivations.

\begin{lemma}
Let $a,b$ be positive integers, and let $s,t$ be tensor shapes. The following statements hold.
\begin{enumerate}
    \item \textbf{(Function composition).} If $f: \mathbb{R}^a \to \mathbb{T}(b)$ and $g: \mathbb{R}^b \to \mathbb{T}(s)$, then $g \circ f: \mathbb{R}^a \to \mathbb{T}(s)$.
    \item \textbf{(Tensor composition).} If $f \in \mathbb{T}(t,b)$ and $g \in \mathbb{T}(b,s)$, then $g \circ f \in \mathbb{T}(t,s)$.
    \item \textbf{(Gradient).} If $f: \mathbb{R}^a \to \mathbb{T}(s)$, then $\nabla f: \mathbb{R}^a \to \mathbb{T}(a,s)$.
    \item \textbf{(Chain rule).} If $f: \mathbb{R}^a \to \mathbb{T}(b)$ and $g: \mathbb{R}^b \to \mathbb{T}(s)$, then $h = g \circ f : \mathbb{R}^a \to \mathbb{T}(s)$ and
    \begin{align*}
    \forall {\boldsymbol{x}} \in \mathbb{R}^a,\qquad
    \nabla h({\boldsymbol{x}}) = (\nabla g \circ f)({\boldsymbol{x}}) \circ \nabla f({\boldsymbol{x}}).
    \end{align*}
    Moreover, if $f,g$ are both tensors (i.e., $f \in \mathbb{T}(a,b)$ and $g \in \mathbb{T}(b,s)$), then $h$ is also a tensor.
    \item \textbf{(Product rule).} If $f: \mathbb{R}^a \to \mathbb{T}(t,b)$ and $g: \mathbb{R}^a \to \mathbb{T}(b,s)$, define $h: \mathbb{R}^a \to \mathbb{T}(t,s)$ by
    $\forall {\boldsymbol{x}} \in \mathbb{R}^a,\ h({\boldsymbol{x}}) = g({\boldsymbol{x}}) \circ f({\boldsymbol{x}})$.
    Then
    \begin{align*}
    \forall {\boldsymbol{x}} \in \mathbb{R}^a,\qquad
    \nabla h({\boldsymbol{x}}) = \nabla g({\boldsymbol{x}}) \circ_2 f({\boldsymbol{x}}) + g({\boldsymbol{x}}) \circ \nabla f({\boldsymbol{x}}).
    \end{align*}
\end{enumerate}
\end{lemma}

\subsection{Proof of the Main Theorems}\label{sec:proof-of-main}

In this subsection, we prove \Cref{thm:main} and \Cref{thm:main-discrete} together. Throughout, we assume that $({\boldsymbol{\theta}},{\boldsymbol{\lambda}})$ is at a good position.

Taking derivatives with respect to ${\boldsymbol{\theta}}$ on both sides of \cref{eq:equivariance}, we obtain
\begin{align}
\nabla f\!\left(H({\boldsymbol{\theta}},{\boldsymbol{\lambda}})\right) \circ \nabla_\theta H({\boldsymbol{\theta}},\lambda)
= \nabla_y G(f({\boldsymbol{\theta}}),{\boldsymbol{\lambda}}) \circ \nabla f({\boldsymbol{\theta}}).
\label{eq:proof-ii}
\end{align}
Taking derivatives with respect to ${\boldsymbol{\lambda}}$ on both sides of \cref{eq:equivariance}, we obtain
\begin{align}
\nabla f\left(H({\boldsymbol{\theta}},{\boldsymbol{\lambda}})\right) \circ \nabla_\lambda H({\boldsymbol{\theta}},{\boldsymbol{\lambda}})
= \nabla_\lambda G(f({\boldsymbol{\theta}}),{\boldsymbol{\lambda}}).
\label{eq:proof-iii}
\end{align}

Since $({\boldsymbol{\theta}},{\boldsymbol{\lambda}})$ is a good position, $\nabla_\theta H({\boldsymbol{\theta}},{\boldsymbol{\lambda}})$ and $\nabla_y G(f({\boldsymbol{\theta}}),{\boldsymbol{\lambda}})$ are invertible. Note that $\nabla_\theta H({\boldsymbol{\theta}},{\boldsymbol{\lambda}})^{-1} \in \mathbb{T}(d,d)$ and $\nabla_y G(f({\boldsymbol{\theta}}),{\boldsymbol{\lambda}})^{-1} \in \mathbb{T}(c,c)$. Therefore, using \cref{eq:proof-ii} and \cref{eq:proof-iii}, we have
\begin{align}
\nabla_y G(f({\boldsymbol{\theta}}), {\boldsymbol{\lambda}}) \circ \nabla f({\boldsymbol{\theta}}) \circ \nabla_\theta H({\boldsymbol{\theta}},{\boldsymbol{\lambda}})^{-1} \circ \nabla_\lambda H({\boldsymbol{\theta}}, {\boldsymbol{\lambda}})
&= \nabla f\!\left(H({\boldsymbol{\theta}}, {\boldsymbol{\lambda}})\right) \circ \nabla_\theta H({\boldsymbol{\theta}}, {\boldsymbol{\lambda}}) \circ \nabla_\theta H({\boldsymbol{\theta}},{\boldsymbol{\lambda}})^{-1} \circ \nabla_\lambda H({\boldsymbol{\theta}}, {\boldsymbol{\lambda}}) \notag
\\
&= \nabla f\!\left(H({\boldsymbol{\theta}}, {\boldsymbol{\lambda}})\right)\circ \nabla_\lambda H({\boldsymbol{\theta}}, {\boldsymbol{\lambda}}) \notag
\\
&= \nabla_\lambda G(f({\boldsymbol{\theta}}), {\boldsymbol{\lambda}}).
\label{eq:proof-iii-p}
\end{align}
Left-multiplying by $\nabla_y G(f({\boldsymbol{\theta}}),{\boldsymbol{\lambda}})^{-1}$, we obtain
\begin{align}
\nabla f({\boldsymbol{\theta}}) \circ \nabla_\theta H({\boldsymbol{\theta}},{\boldsymbol{\lambda}})^{-1} \circ \nabla_\lambda H({\boldsymbol{\theta}}, {\boldsymbol{\lambda}})
= \nabla_y G(f({\boldsymbol{\theta}}), {\boldsymbol{\lambda}})^{-1} \circ \nabla_\lambda G(f({\boldsymbol{\theta}}), {\boldsymbol{\lambda}}).
\end{align}
Using $\nabla L({\boldsymbol{\theta}}) = \nabla \ell(f({\boldsymbol{\theta}})) \circ \nabla f({\boldsymbol{\theta}})$, we obtain \cref{eq:main-first}.

\paragraph{Second-order identities.}

Now, right-composing \cref{eq:proof-ii} with $\nabla_\theta H({\boldsymbol{\theta}}, {\boldsymbol{\lambda}})^{-1}$, we obtain
\begin{align}
\nabla f\left(H({\boldsymbol{\theta}}, {\boldsymbol{\lambda}})\right)
= \nabla_y G(f({\boldsymbol{\theta}}), {\boldsymbol{\lambda}}) \circ \nabla f({\boldsymbol{\theta}})\circ \nabla_\theta H({\boldsymbol{\theta}}, {\boldsymbol{\lambda}})^{-1}.
\label{eq:proof-ii-p}
\end{align}

Taking derivatives with respect to ${\boldsymbol{\theta}}$ on both sides of \cref{eq:proof-ii}, we obtain
\begin{align}
& \nabla^2 f \left(H({\boldsymbol{\theta}}, {\boldsymbol{\lambda}})\right) \circ \nabla_\theta H({\boldsymbol{\theta}}, {\boldsymbol{\lambda}})  \circ_{\pi} \nabla_\theta H({\boldsymbol{\theta}}, {\boldsymbol{\lambda}})
+ \nabla f(H({\boldsymbol{\theta}},{\boldsymbol{\lambda}})) \circ \nabla_\theta^2 H({\boldsymbol{\theta}},{\boldsymbol{\lambda}}) \notag
\\
=\, & \nabla_y^2 G(f({\boldsymbol{\theta}}), {\boldsymbol{\lambda}}) \circ \nabla f({\boldsymbol{\theta}}) \circ_2 \nabla f({\boldsymbol{\theta}})
+ \nabla_y G (f({\boldsymbol{\theta}}), \lambda) \circ \nabla^2 f({\boldsymbol{\theta}}).
\label{eq:proof-iv}
\end{align}
Taking derivatives with respect to ${\boldsymbol{\lambda}}$ on both sides of \cref{eq:proof-ii}, we obtain
\begin{align}
&\nabla^2 f \left(H({\boldsymbol{\theta}}, {\boldsymbol{\lambda}})\right) \circ \nabla_\lambda H({\boldsymbol{\theta}}, {\boldsymbol{\lambda}})  \circ_{\pi} \nabla_\theta H({\boldsymbol{\theta}}, {\boldsymbol{\lambda}})
+ \nabla f(H({\boldsymbol{\theta}},{\boldsymbol{\lambda}})) \circ \nabla_\lambda \nabla_\theta H({\boldsymbol{\theta}},{\boldsymbol{\lambda}}) \notag
\\
=\, &  \nabla_\lambda \nabla_y G(f({\boldsymbol{\theta}}), {\boldsymbol{\lambda}}) \circ_2 \nabla f({\boldsymbol{\theta}}).
\label{eq:proof-v}
\end{align}
Taking derivatives with respect to ${\boldsymbol{\lambda}}$ on both sides of \cref{eq:proof-iii}, we obtain
\begin{align}
\nabla^2 f \left(H({\boldsymbol{\theta}}, {\boldsymbol{\lambda}})\right) \circ \nabla_\lambda H({\boldsymbol{\theta}}, {\boldsymbol{\lambda}}) \circ_{\pi} \nabla_\lambda H({\boldsymbol{\theta}}, {\boldsymbol{\lambda}})
+ \nabla f(H({\boldsymbol{\theta}},{\boldsymbol{\lambda}})) \circ \nabla_\lambda^2 H({\boldsymbol{\theta}},{\boldsymbol{\lambda}})
= \nabla_{\lambda}^2 G (f({\boldsymbol{\theta}}), \lambda).
\label{eq:proof-vi}
\end{align}

For simplicity, we denote $A = X({\boldsymbol{\theta}}, {\boldsymbol{\lambda}})$ and $B = Y({\boldsymbol{\theta}}, {\boldsymbol{\lambda}})$. Right-composing \cref{eq:proof-iv} with $A$ and using \cref{eq:main-first}, we get
\begin{align}
& \nabla^2 f \left(H({\boldsymbol{\theta}}, {\boldsymbol{\lambda}})\right) \circ \nabla_\theta H({\boldsymbol{\theta}}, {\boldsymbol{\lambda}})  \circ_{\pi} \nabla_\theta H({\boldsymbol{\theta}}, {\boldsymbol{\lambda}})\circ A
+ \nabla f(H({\boldsymbol{\theta}},{\boldsymbol{\lambda}})) \circ \nabla_\theta^2 H({\boldsymbol{\theta}},{\boldsymbol{\lambda}}) \circ A \notag
\\
=\, & \nabla_y^2 G(f({\boldsymbol{\theta}}), {\boldsymbol{\lambda}}) \circ (\nabla f({\boldsymbol{\theta}}) \circ A) \circ_2 \nabla f({\boldsymbol{\theta}})
+ \nabla_y G (f({\boldsymbol{\theta}}), \lambda) \circ \nabla^2 f({\boldsymbol{\theta}}) \circ A \notag
\\
=\, & \nabla_y^2 G(f({\boldsymbol{\theta}}), {\boldsymbol{\lambda}}) \circ B \circ_2 \nabla f({\boldsymbol{\theta}})
+ \nabla_y G (f({\boldsymbol{\theta}}), \lambda) \circ \nabla^2 f({\boldsymbol{\theta}}) \circ A.
\label{eq:proof-iv-p}
\end{align}
Note that the first term on the left-hand side of \cref{eq:proof-iv-p} is the same as the first term on the left-hand side of \cref{eq:proof-v}. Therefore, we can use \cref{eq:proof-v} to eliminate this term. Moreover, we can use \cref{eq:proof-ii-p} to eliminate $\nabla f(H({\boldsymbol{\theta}},{\boldsymbol{\lambda}}))$. We obtain
\begin{align}
\nabla^2 f({\boldsymbol{\theta}}) \circ A
=\, &  - \nabla f({\boldsymbol{\theta}})\circ \nabla_\theta H({\boldsymbol{\theta}}, {\boldsymbol{\lambda}}) ^{-1} \circ \nabla_\lambda \nabla_\theta H({\boldsymbol{\theta}},{\boldsymbol{\lambda}}) \notag
\\
& + \nabla f({\boldsymbol{\theta}})\circ \nabla_\theta H({\boldsymbol{\theta}}, {\boldsymbol{\lambda}}) ^{-1} \circ \nabla_\theta^2 H({\boldsymbol{\theta}},{\boldsymbol{\lambda}}) \circ A \notag
\\
& + \nabla_y G(f({\boldsymbol{\theta}}), {\boldsymbol{\lambda}})^{-1} \circ \nabla_\lambda \nabla_y G(f({\boldsymbol{\theta}}), {\boldsymbol{\lambda}}) \circ_2 \nabla f({\boldsymbol{\theta}}) \notag
\\
& - \nabla_y G(f({\boldsymbol{\theta}}), {\boldsymbol{\lambda}})^{-1} \circ \nabla_y^2 G(f({\boldsymbol{\theta}}), {\boldsymbol{\lambda}}) \circ B \circ_2 \nabla f({\boldsymbol{\theta}}).
\label{eq:proof-star2}
\end{align}

Similarly, if we right-compose \cref{eq:proof-iv} with $A$ in its first two arguments (i.e., compose with $A \circ_2 A$), we get
\begin{align}
& \nabla^2 f \left(H({\boldsymbol{\theta}}, {\boldsymbol{\lambda}})\right) \circ \nabla_\lambda H({\boldsymbol{\theta}}, {\boldsymbol{\lambda}})  \circ_{\pi} \nabla_\lambda H({\boldsymbol{\theta}}, {\boldsymbol{\lambda}})
+ \nabla f(H({\boldsymbol{\theta}},{\boldsymbol{\lambda}})) \circ \nabla_\theta^2 H({\boldsymbol{\theta}},{\boldsymbol{\lambda}}) \circ A \circ_2 A \notag
\\
=\, & \nabla_y^2 G(f({\boldsymbol{\theta}}), {\boldsymbol{\lambda}}) \circ B \circ_2 B
+ \nabla_y G (f({\boldsymbol{\theta}}), \lambda) \circ \nabla^2 f({\boldsymbol{\theta}}) \circ A \circ_2 A.
\label{eq:proof-v-p}
\end{align}
Using \cref{eq:proof-vi} and \cref{eq:proof-ii-p} in \cref{eq:proof-v-p}, we obtain
\begin{align}
\nabla^2 f({\boldsymbol{\theta}}) \circ A \circ_2 A
=\, &  -  \nabla f({\boldsymbol{\theta}})\circ \nabla_\theta H({\boldsymbol{\theta}}, {\boldsymbol{\lambda}}) ^{-1} \circ \nabla_\lambda^2 H({\boldsymbol{\theta}},{\boldsymbol{\lambda}}) \notag
\\
& + \nabla f({\boldsymbol{\theta}})\circ \nabla_\theta H({\boldsymbol{\theta}}, {\boldsymbol{\lambda}}) ^{-1} \circ \nabla_\theta^2 H({\boldsymbol{\theta}},{\boldsymbol{\lambda}}) \circ A \circ_2 A \notag
\\
& + \nabla_y G(f({\boldsymbol{\theta}}), {\boldsymbol{\lambda}})^{-1} \circ  \nabla_{\lambda}^2 G (f({\boldsymbol{\theta}}), \lambda) \notag
\\
& - \nabla_y G(f({\boldsymbol{\theta}}), {\boldsymbol{\lambda}})^{-1} \circ \nabla_y^2 G(f({\boldsymbol{\theta}}), {\boldsymbol{\lambda}}) \circ B \circ_2 B.
\label{eq:proof-star3}
\end{align}

Using $\nabla L(\theta) = \nabla \ell( f({\boldsymbol{\theta}})) \circ \nabla f({\boldsymbol{\theta}})$ and
\begin{align}\label{eq:hessian-model-to-loss}
\nabla^2 L(\theta) = \nabla^2 \ell(f({\boldsymbol{\theta}})) \circ \nabla f({\boldsymbol{\theta}}) \circ_2 \nabla f({\boldsymbol{\theta}}) + \nabla \ell(f({\boldsymbol{\theta}})) \circ \nabla^2 f({\boldsymbol{\theta}}),
\end{align}
in \cref{eq:proof-star2} and \cref{eq:proof-star3}, we obtain \cref{eq:main-second-1} and \cref{eq:main-second-2}, respectively.

\paragraph{Discrete identities.}
In the discrete case, taking derivatives with respect to ${\boldsymbol{\lambda}}$ is prohibited, since we do not have structural information about $H$ with respect to ${\boldsymbol{\lambda}}$. Therefore, only \cref{eq:proof-ii} and \cref{eq:proof-iv} are valid. For a specific ${\boldsymbol{\lambda}}_0 \in \mathbb{R}^p$, if ${\boldsymbol{\theta}} \in \mathop{\mathrm{Fix}}(H, {\boldsymbol{\lambda}}_0)$, then $H({\boldsymbol{\theta}}, {\boldsymbol{\lambda}}_0) = {\boldsymbol{\theta}}$. Substituting this identity into \cref{eq:proof-ii} and \cref{eq:proof-iv}, we obtain
\begin{align}
\nabla f({\boldsymbol{\theta}}) \circ \widehat A =  \widehat B \circ \nabla f({\boldsymbol{\theta}}),
\label{eq:proof-dis-1}
\end{align}
and
\begin{align}
\nabla^2 f (\theta) \circ \widehat A  \circ_{\pi} \widehat A + \nabla f({\boldsymbol{\theta}}) \circ \nabla_\theta^2 H({\boldsymbol{\theta}},{\boldsymbol{\lambda}}_0)
= \nabla_y^2 G(f({\boldsymbol{\theta}}), {\boldsymbol{\lambda}}_0) \circ \nabla f({\boldsymbol{\theta}}) \circ_2 \nabla f({\boldsymbol{\theta}})  + \widehat B \circ \nabla^2 f({\boldsymbol{\theta}}),
\label{eq:proof-dis-2}
\end{align}
where $\widehat A = \nabla_\theta H({\boldsymbol{\theta}}, {\boldsymbol{\lambda}}_0)$ and $\widehat B = \nabla_y G(f({\boldsymbol{\theta}}), {\boldsymbol{\lambda}}_0)$. Since we only consider symmetry here, i.e., $G = \mathrm{Id}$, we have $\widehat B = \mathrm{Id}$. Using $\nabla L({\boldsymbol{\theta}}) = \nabla \ell( f({\boldsymbol{\theta}})) \circ \nabla f({\boldsymbol{\theta}})$ and \cref{eq:hessian-model-to-loss} in \cref{eq:proof-dis-1} and \cref{eq:proof-dis-2}, we obtain \cref{eq:discrete-first} and \cref{eq:dis-realization-second}, respectively.

\subsection{Proof of \Cref{cor:gradient-hessian-eigen-decomp}}

If $\ell'(y) = 0$ or $my\ell''(y) + (m-1)\ell'(y) = 0$, then $\nabla^2 L({\boldsymbol{\theta}})\,{\boldsymbol{\theta}} = {\boldsymbol{0}}$, in which case the proposition clearly holds. In the following, we assume $\ell'(y) \neq 0$ and $my\ell''(y) + (m-1)\ell'(y) \neq 0$.

Let $\alpha({\boldsymbol{\theta}}) = \left(my\frac{\ell''(y)}{\ell'(y)} + m-1\right)^{-1}$. Since $\ell'(y) \neq 0$ and $my\ell''(y) + (m-1)\ell'(y) \neq 0$, $\alpha$ is well-defined. Using \cref{eq:realization-second-homo-1}, we have
\begin{align*}
\left< {\boldsymbol{g}}, {\boldsymbol{u}}_k\right>
&= \alpha({\boldsymbol{\theta}}) \left<{\boldsymbol{u}}_k, \nabla^2 L({\boldsymbol{\theta}})\, {\boldsymbol{\theta}}\right>
\\
&= \alpha({\boldsymbol{\theta}})\left<\nabla^2 L({\boldsymbol{\theta}})\,{\boldsymbol{u}}_k, {\boldsymbol{\theta}}\right>
\\
&= \alpha({\boldsymbol{\theta}})\lambda_k\left<  {\boldsymbol{u}}_k, {\boldsymbol{\theta}}\right>.
\end{align*}

\subsection{Proof of \Cref{cor:neother-flow}}

Using the characterization in \citep{ziyin_sgd}, Section~4.1, we have
\begin{align*}
\dot C({\boldsymbol{\theta}}) = \sigma^2 \mathrm{Tr}\!\left(\Sigma({\boldsymbol{\theta}})\, \nabla^2 C({\boldsymbol{\theta}})\right).
\end{align*}

Note that $\nabla^2 C({\boldsymbol{\theta}}) = \nabla_{{\theta}} \nabla_\lambda H({\boldsymbol{\theta}},0)$. Since the symmetry holds for every input datum $x$, by \Cref{thm:main} and \cref{eq:main-second-1}, we have
\begin{align*}
\nabla^2 C({\boldsymbol{\theta}})\,\nabla L_x({\boldsymbol{\theta}})
= -\nabla^2 L_x({\boldsymbol{\theta}})\,\nabla C({\boldsymbol{\theta}}).
\end{align*}
Moreover, the symmetry also holds for the expected loss (since it does not change the model output); therefore, we also have
\begin{align*}
\nabla^2 C({\boldsymbol{\theta}})\,\nabla L({\boldsymbol{\theta}})
= - \nabla^2 L({\boldsymbol{\theta}})\,\nabla C({\boldsymbol{\theta}}).
\end{align*}

Therefore,
\begin{align*}
\frac{1}{\sigma^2}\dot C({\boldsymbol{\theta}})
&= \mathrm{Tr}\!\left(\Sigma({\boldsymbol{\theta}})\, \nabla^2 C({\boldsymbol{\theta}})\right)
\\
&= \mathbb{E}\,\mathrm{Tr}\!\left(\nabla L_x({\boldsymbol{\theta}})\, \nabla L_x({\boldsymbol{\theta}})^\top \nabla^2 C({\boldsymbol{\theta}})\right)
- \mathrm{Tr}\!\left(\nabla L({\boldsymbol{\theta}})\, \nabla L({\boldsymbol{\theta}})^\top \nabla^2 C({\boldsymbol{\theta}})\right)
\\
&= \mathbb{E}\left<\nabla L_x({\boldsymbol{\theta}}), \nabla^2 C({\boldsymbol{\theta}})\, \nabla L_x({\boldsymbol{\theta}})\right>
- \left<\nabla L({\boldsymbol{\theta}}), \nabla^2 C({\boldsymbol{\theta}})\, \nabla L({\boldsymbol{\theta}})\right>
\\
&= -\mathbb{E}\left<\nabla L_x({\boldsymbol{\theta}}), \nabla^2 L_x({\boldsymbol{\theta}})\, \nabla C({\boldsymbol{\theta}})\right>
+ \left<\nabla L({\boldsymbol{\theta}}), \nabla^2 L({\boldsymbol{\theta}})\, \nabla C({\boldsymbol{\theta}})\right>
\\
&= -\mathbb{E}\left<\nabla^2 L_x({\boldsymbol{\theta}})\nabla L_x({\boldsymbol{\theta}}),  \nabla C({\boldsymbol{\theta}})\right>
+ \left<\nabla^2 L({\boldsymbol{\theta}})\nabla L({\boldsymbol{\theta}}),  \nabla C({\boldsymbol{\theta}})\right>
\\
&= -\left<\nabla C({\boldsymbol{\theta}}), \mathbb{E}\,\nabla^2 L_x({\boldsymbol{\theta}})\nabla L_x({\boldsymbol{\theta}})
- \nabla^2 L({\boldsymbol{\theta}})\nabla L({\boldsymbol{\theta}})\right>
\\
&= \frac 12\left<\nabla C({\boldsymbol{\theta}}) , \frac{\partial }{\partial {\boldsymbol{\theta}}}\mathrm{Tr}\!\left(\Sigma({\boldsymbol{\theta}})\right)\right>.
\end{align*}

\subsection{Proof of \Cref{cor:last-layer-alignment}}

Substituting the setting into \Cref{thm:main} \cref{eq:main-second-2}, we obtain that for any matrix ${\boldsymbol{U}} \in \mathbb{R}^{c\times c}$,
\begin{align*}
\left\langle \mathop{ \mathrm{ vec } }({\boldsymbol{U}}{\boldsymbol{W}}),\nabla_{\mathop{ \mathrm{ vec } }({\boldsymbol{W}})}^2 L({\boldsymbol{\theta}})\,\mathop{ \mathrm{ vec } }({\boldsymbol{U}}{\boldsymbol{W}})\right\rangle
&=\left\langle {\boldsymbol{U}}{\boldsymbol{y}},\,\nabla^2\ell({\boldsymbol{y}})\,{\boldsymbol{U}}{\boldsymbol{y}}\right\rangle.
\end{align*}
Let ${\boldsymbol{V}} = {\boldsymbol{U}} {\boldsymbol{W}}$. Since ${\boldsymbol{U}}$ ranges over all $c\times c$ matrices, ${\boldsymbol{V}}$ ranges over all $c\times s$ matrices whose row space is contained in the row space of ${\boldsymbol{W}}$. Note also that ${\boldsymbol{y}} = {\boldsymbol{W}}{\boldsymbol{h}}$. Therefore,
\begin{align*}
\left< \mathop{ \mathrm{ vec } }({\boldsymbol{V}}),\nabla_{\mathop{ \mathrm{ vec } }({\boldsymbol{W}})}^2 L({\boldsymbol{\theta}})\,\mathop{ \mathrm{ vec } }({\boldsymbol{V}})\right>
=\left< {\boldsymbol{U}} {\boldsymbol{W}}{\boldsymbol{h}},\,\nabla^2\ell({\boldsymbol{y}})\,{\boldsymbol{U}}{\boldsymbol{W}}{\boldsymbol{h}}\right>
= \left< {\boldsymbol{V}} {\boldsymbol{h}},\,\nabla^2\ell({\boldsymbol{y}})\,{\boldsymbol{V}}{\boldsymbol{h}}\right>.
\end{align*}

\subsection{Proof of \Cref{cor:mirro-sym}}

First, note that \cref{eq:dis-realization-first} implies ${\boldsymbol{O}}{\boldsymbol{O}}^\top \nabla L({\boldsymbol{\theta}}) = {\boldsymbol{0}}$, which is equivalent to ${\boldsymbol{O}}^\top \nabla L({\boldsymbol{\theta}}) = {\boldsymbol{0}}$.

Moreover, \cref{eq:dis-realization-second} implies
\begin{align*}
{\boldsymbol{P}}\nabla^2 L({\boldsymbol{\theta}}) = \nabla^2 L({\boldsymbol{\theta}}){\boldsymbol{P}}.
\end{align*}
Therefore, for any ${\boldsymbol{x}} \in \mathop{ \mathrm{ col } }({\boldsymbol{O}})$,
\begin{align*}
{\boldsymbol{P}} \nabla ^2L({\boldsymbol{\theta}}) {\boldsymbol{x}}
= \nabla^2 L({\boldsymbol{\theta}}) {\boldsymbol{P}} {\boldsymbol{x}}
= -\nabla ^2 L({\boldsymbol{\theta}}) {\boldsymbol{x}},
\end{align*}
which means $\nabla ^2 L({\boldsymbol{\theta}}){\boldsymbol{x}}$ is an eigenvector of ${\boldsymbol{P}}$ with eigenvalue $-1$, and hence $\nabla ^2 L({\boldsymbol{\theta}}){\boldsymbol{x}} \in \mathop{ \mathrm{ col } }({\boldsymbol{O}})$. Similarly, for ${\boldsymbol{y}} \perp  \mathop{ \mathrm{ col } }({\boldsymbol{O}})$, we have
\begin{align*}
{\boldsymbol{P}} \nabla ^2L({\boldsymbol{\theta}}) {\boldsymbol{y}}
= \nabla^2 L({\boldsymbol{\theta}}) {\boldsymbol{P}} {\boldsymbol{y}}
= \nabla ^2 L({\boldsymbol{\theta}}) {\boldsymbol{y}},
\end{align*}
which means $\nabla ^2 L({\boldsymbol{\theta}}){\boldsymbol{y}}$ lies in the eigenspace of ${\boldsymbol{P}}$ corresponding to eigenvalue $1$. This is equivalent to $\nabla ^2 L({\boldsymbol{\theta}}) {\boldsymbol{y}} \perp \mathop{ \mathrm{ col } }({\boldsymbol{O}})$.